\documentclass[letterpaper, 10 pt, conference]{ieeeconf}

\usepackage{amsmath,amsfonts}
\usepackage{algorithmic}
\usepackage{array}
\usepackage{textcomp}
\usepackage{stfloats}
\usepackage{url}
\usepackage{verbatim}
\usepackage{graphicx}
\usepackage{balance}

\usepackage{bm}
\usepackage{todonotes}
\usepackage{subcaption}

\usepackage{amssymb}  

\makeatletter
\let\NAT@parse\undefined
\makeatother
\usepackage{hyperref}  
\usepackage{cleveref}

\usepackage{listings}
\IEEEoverridecommandlockouts                              
\title{\LARGE \bf Stabilizing Humanoid Robot Trajectory Generation via \\
Physics-Informed Learning and Control-Informed Steering}
\author{Evelyn D'Elia$^{1,2}$, Paolo Maria Viceconte$^{1}$, Lorenzo Rapetti$^{1}$, Diego Ferigo$^{1}$, Giulio Romualdi$^{1}$, \\ Giuseppe L'Erario$^{1,2}$, Raffaello Camoriano$^{3,4}$, and Daniele Pucci$^{1,2}$
\thanks{*This work was supported by the ergoCub project.\newline
RC acknowledges the following: this study was carried out within the FAIR - Future Artificial Intelligence Research and received funding from the European Union Next-GenerationEU (PIANO NAZIONALE DI RIPRESA E RESILIENZA (PNRR) – MISSIONE 4 COMPONENTE 2, INVESTIMENTO 1.3 – D.D. 1555 11/10/2022, PE00000013). This manuscript reflects only the authors’ views and opinions, neither the European Union nor the European Commission can be considered responsible for them. }
\thanks{$^{1}$Artificial and Mechanical Intelligence, Italian Institute of Technology, 16163 Genoa, Italy
        {\tt\small evelyn.delia@iit.it}}%
\thanks{$^{2}$Machine Learning and Optimisation, University of Manchester, M13 9PL Manchester, U.K.}%
\thanks{$^{3}$Dipartimento di Automatica e Informatica, Politecnico di Torino, 10129 Turin, Italy}%
\thanks{$^{4}$Rehab Technologies Lab, Italian Institute of Technology, 16163 Genoa, Italy}%
}

\begin{document}

\maketitle

\begin{abstract}
Recent trends in humanoid robot control have successfully employed imitation learning to
enable the learned generation of smooth, human-like trajectories
from 
human data. 
While these approaches make more realistic motions possible, they are limited by the amount of available motion data, and do not incorporate prior knowledge about the physical laws governing the system and its interactions with the environment.
Thus they may violate
such laws, leading to divergent trajectories and sliding contacts which limit real-world stability.
We 
address such limitations via a two-pronged learning strategy which leverages the
known physics
of the 
system and 
fundamental
control principles. 
First, 
we encode physics priors during supervised imitation learning
to promote 
trajectory feasibility.
Second, we minimize drift at inference time
by applying a proportional-integral controller directly to the generated output state.
We 
validate
our method 
on various locomotion behaviors for the ergoCub humanoid robot, where a physics-informed loss
encourages zero contact foot velocity.
Our 
experiments
demonstrate that 
the proposed approach
is compatible with
multiple
controllers on a real robot
and
significantly 
improves the accuracy and physical constraint conformity of generated trajectories.

\end{abstract}



\section{Introduction}
When working in collaborative, human-centric environments, humanoid robots need to perform
in a smooth, safe fashion.
Moreover, in collaborative scenarios it is important for the robot's motions to be \emph{human-like}, 
such that the human can trust, understand, and predict its behavior \cite{Santis2008}, \cite{Riek2009}, as well as to improve the robot's performance in tasks designed for humans. 
However, human-likeness is
not a clearly defined property, and enforcing it in trajectory generation by explicit modeling is impractical due to high computational cost thanks to the highly nonlinear dynamics of a humanoid robot.

Human-like motion cannot be achieved without a control architecture that is able to stabilize the desired robot behavior. 
This is challenging due to the large number of degrees of freedom 
and complex stability constraints.
Many state-of-the-art approaches exploit a hierarchical structure to address
the problem: \emph{trajectory optimization}, which generates desired trajectories, \emph{trajectory adjustment}, which stabilizes the trajectory, and \emph{trajectory control}, which solves a quadratic programming (QP) problem to translate the trajectory into motor references~\cite{Romualdi2018}.

Humanoid trajectory optimization is categorized as either \emph{model-based} or \emph{model-free}. \emph{Model-based} techniques, which employ an explicit model of the system, may describe the full dynamics or a simplified/reduced version. Full dynamics approaches achieve high-fidelity motions,
but require meticulous design and are computationally expensive.
Simplified model-based approaches, such as representing locomotion with an inverted pendulum model \cite{Kajita}, or centroidal dynamics models \cite{Englsberger2015}, \cite{Dai2014}, 
adopt a low-dimensional parameterization of the complex motion of a humanoid to improve optimization efficiency.
However, the use of simplified models for trajectory generation hinders their human-likeness.

\begin{figure}
    \centering
    \includegraphics[width=\linewidth]{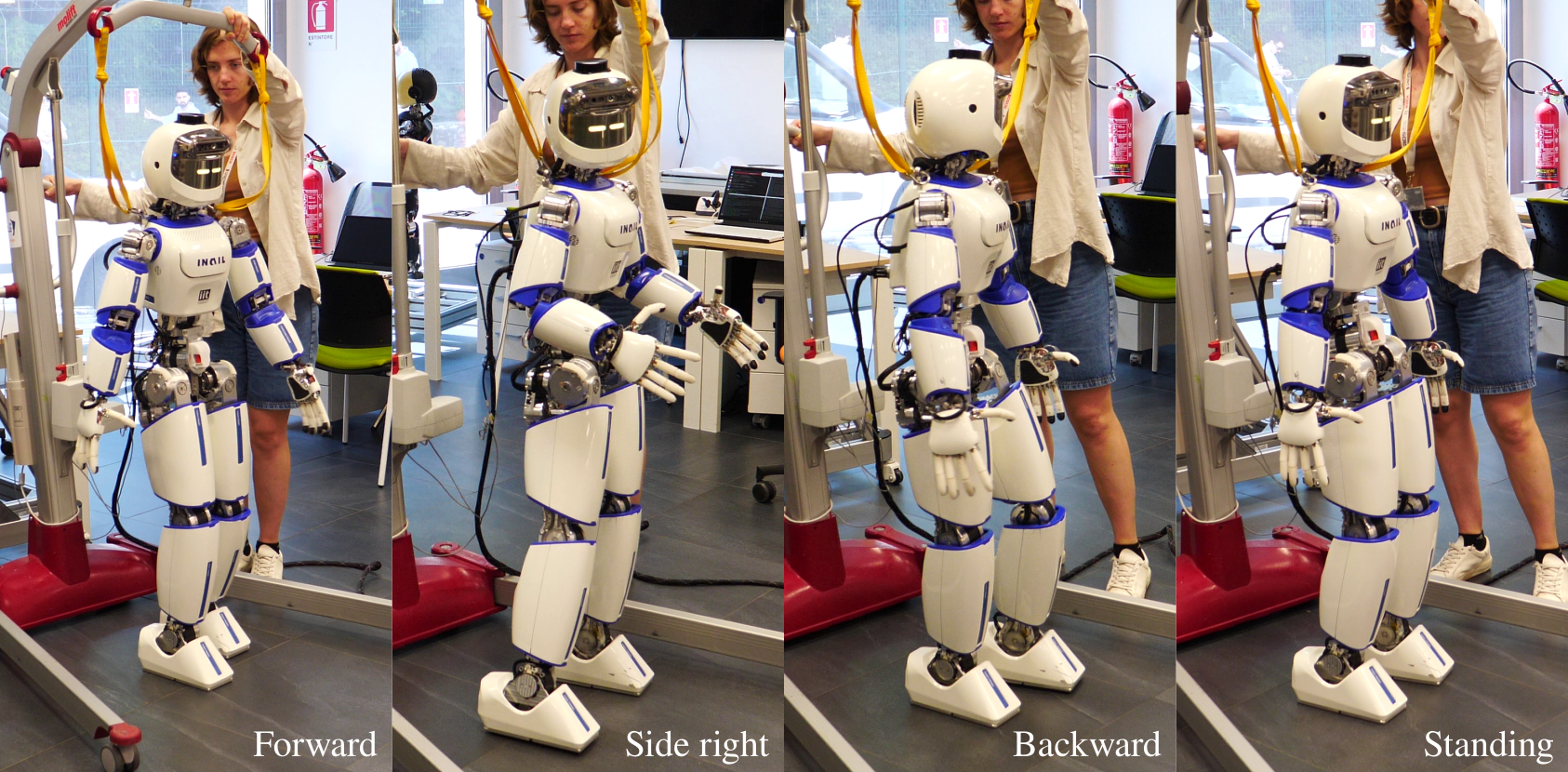}
    \caption{Our method used to execute various walking directions in a single trajectory on the ergoCub robot.}
    \label{fig:sn000_walking}
\end{figure}

Alternatively, \emph{model-free} approaches use optimization methods to directly approximate training data without the use of an explicit model, allowing them to achieve better computational efficiency than model-based methods.
One class of trajectory generation solutions which reduce model dependence is \emph{imitation learning (IL)}. One special case of IL, which relies entirely on mimicking the dataset, is called \emph{behavior cloning (BC)} and has been applied to achieve legged locomotion trajectory planning in simulation \cite{Holden2017, Tsounis2020, Zhang2018} and more recently on real robots \cite{Viceconte2022, Huang2025} by harnessing task-specific training data.
IL has also been employed to supplement the reward for learning human-like policies with reinforcement learning (RL) \cite{Peng2021,Peng2022,Hassan2023,Tang2023, Serifi2024}.
In recent years, the use of IL in an RL context has led to robust real-world humanoid control \cite{Dao2023, Gu2024, Radosavovic2024}.

The advantages of IL over RL include a much better sample efficiency, a simpler supervised learning problem, and reduced computational requirements.
However, IL is limited by the quantity and quality of available demonstration data~\cite{Karniadakis2021}.
As a result, trajectories generated via IL can break physical laws and exhibit asymmetry or bias toward certain modes.
To overcome this weakness, 
in recent years 
\emph{physics-informed machine learning (PIML)} 
has been proposed \cite{Raissi2019}. 
PIML 
incorporates priors about the physical system dynamics into the learning process,
thus combining the strengths of model-based 
and learning-based methods.
A key strength of PIML lies in its ability to encode \emph{differentiable} 
priors
during learning, which can speed up convergence and reduce the amount of data needed to infer the underlying physical laws.
This strategy is rendered computationally feasible by
automatic differentiation (AD),
which is nowadays available in most machine learning libraries.

PIML proved effective at accurately learning to emulate partial differential equations (PDEs) to model complex physical systems~\cite{Lusch2018,Gin2020}. 
In the context of learning for humanoid control, incorporating physics priors as part of the reward function has been shown in simulation to counteract learned artifacts related to ground contacts in humanoid trajectory generation~\cite{Yuan2023}. 
However, the physical reward components in~\cite{Yuan2023} are not differentiable and thus cannot influence backpropagation.
Additionally, smooth humanoid locomotion has been achieved on real-world robots with PIML by formulating a Lipschitz constraint as a reward function \cite{chen2024lcp}.
However, this approach does not use IL and the resulting policies are less human-like.

In summary, BC methods have achieved notable success in trajectory generation and control for humanoids in recent years. In parallel, PIML has been shown to improve physical fidelity when training examples are scarce or unevenly distributed. Therefore, the incorporation of PIML into BC-based humanoid locomotion offers significant potential to improve the stability of generated trajectories while preserving the human-likeness learned from the data.


In this work, we propose a modular PIML and control theory-inspired framework to improve physical fidelity and mitigate learned bias of BC for a humanoid robot.
First, we introduce a supervised learning approach to predict the full next state of the robot given the previous state and a user-provided desired direction input.
We then propose improvements which enhance the symmetry and physical fidelity of the final trajectory.
We emphasize that our contributions are modular.
%
Our contributions are summarized as follows:
\begin{itemize}
    \item We present a framework
    to learn the full kinematic state of a floating-base robot while considering kinematically feasible contacts via 
    a \emph{physics-informed} loss, in addition to a standard data-fitting loss term.
    Specifically, 
    the network learns to keep the contact foot still while the swing foot moves.
    \item We harness fundamental control concepts to impose a corrective
    term on the output of our network. 
    This \emph{control-informed} blending of user input with network predictions reduces the drift of generated trajectories.
    \item We perform ablation studies and real-world validation experiments on the ergoCub robot using two 
    control architectures, highlighting the efficacy and modularity of our approach.
\end{itemize}


This paper is structured as follows: \Cref{sec:background} introduces the relevant background information, \Cref{sec:methods} describes the details of our approach, \Cref{sec:results} contains an analysis and discussion of experimental results, and \Cref{sec:conclusion} summarizes our contribution and proposes future work.

\section{BACKGROUND}
\label{sec:background}
\subsection{Notation}
\begin{itemize}
    \item $\bm{\mathcal F}$ and $\bm{\mathcal{B}}$ denote the nonlinear differential and boundary condition operators for an arbitrary differential equation.
    \item $\bm u(\bm z)$ indicates the differential equation solution as a function of the state and time vector $\bm z := [\bm x_1, \dots, \bm x_d; t]$.
    \item $\bm f(\bm z)$ 
    and $\bm g(\bm z)$ are, respectively, the functions
    defining the data dynamics and boundary conditions.
    \item $\mathcal L(\bm z;\theta)$ is a loss function computed with respect to all unknown parameters $\theta$.
    \item $\mathcal{B}$ and $\mathcal{I}$ are the robot base and inertial frames, where $\mathcal{B}$ is located at the center of the waist with the $x$ axis pointing forward and the $z$ axis pointing upwards.
    \item ${}^{\mathcal{A}} \bm R_{\mathcal{C}}$ is the matrix which describes the rotation from frame $\mathcal{A}$ to frame $\mathcal{C}$.
    \item $\mathbf v = (\boldsymbol v, \bm \omega) \in \mathbb R^6$ is a 6D velocity vector, where $\boldsymbol{v}$ is the linear velocity, $\bm \omega$ is the angular velocity.
    \item $\bm s, \dot {\bm s} \in \mathbb R^n$ are the joint positions and velocities, respectively,  where $n$ is the number of considered joints.
    \item $\bm q = (\bm q_{\mathcal B}, \bm s)\in \mathbb R^{6+n}$ is the combined position vector of the floating base model, where $\bm q_{\mathcal B} = (\bm p_{\mathcal{B}}, \bm \psi_{\mathcal{B}})$ contains the floating base position $\bm p_{\mathcal{B}}$ and extrinsic Tait-Bryan angles $\bm \psi_{\mathcal{B}}$.
    \item $\bm \nu = (\mathbf v_{\mathcal B}, \bm{\dot{s}})\in \mathbb R^{6+n}$ is the combined velocity vector of the floating base model.
    \item $\mathbf J$ is the Jacobian matrix, which is a function of the robot state, and describes the relationship between $\bm \nu$ and $\mathbf v$ at each robot link.
    \item The operator $(\cdot)^{\top}$ takes the transpose of a matrix.
    \item A variable with a hat $\hat{(\cdot)}$ denotes a predicted value.
\end{itemize}

\subsection{Physics-Informed Machine Learning}
PIML is  
 based on 
injecting prior knowledge into the learning process to promote solutions which comply with the physical laws governing the system.
This can be particularly useful for learning high-dimensional systems, especially when the amount of data is limited, 
since PIML can compensate for the curse of dimensionality by relying on known system properties~\cite{Karniadakis2021}. 
As a result, PIML is promising for humanoid control, where the interactions with the environment are high-dimensional, non-linear, and difficult to fully capture in a finite dataset.
Using physics-informed (PI) methods, the gradient of quantities involved in humanoid dynamics, such as the Jacobian, are computed easily with AD, thus allowing them to directly influence learning.
This is in contrast to many traditional RL methods, which may integrate physical laws via reward terms, but do not leverage the gradient.
In this paper, we implement PIML via the loss function, 
exploiting AD to 
compute 
derivatives of the PI loss during backpropagation. 
We define a differential equation and its initial
conditions based on \cite[Sec.~II]{Cuomo2022}:
\begin{equation}
    \begin{aligned}
        \bm {\mathcal F} (\bm u(\bm z); \gamma) &= \bm f(\bm z) \quad \bm z \in \Omega, \\
    \bm {\mathcal B}(\bm u(\bm z)) &= \bm g(\bm z) 
    \quad \bm z \in \partial\Omega,
    \end{aligned}
    \label{eq:pi_formalism}
\end{equation}
where $\gamma$ is the set of physical parameters of the nonlinear differential equation, $\Omega$ is a manifold, and $\partial \Omega$ is the manifold's boundary. 
A
physics-informed neural network (PINN) $\hat {\bm u}_{\theta} (\bm z)$ 
aims to approximate
the solution 
such that
\begin{equation}
    \hat {\bm u}_{\theta} (\bm z) \approx \bm u (\bm z),
\end{equation}
by minimizing a physics-informed loss taking the following form with respect to NN weights $\theta$:
\begin{equation}
        \mathcal L(\bm z;\theta) = w_{\bm{\mathcal F}} \mathcal L_{\bm{\mathcal F}}(\bm z;\theta) + w_{\bm{\mathcal B}} \mathcal L_{\bm{\mathcal B}}(\bm z;\theta) + w_{\bm{\mathcal D}} \mathcal L_{\bm{\mathcal D}}(\bm z;\theta), 
        \label{eq:pi_loss_general}
\end{equation}
where the $\bm{\mathcal F}$ component depends on the differential equation, the $\bm{\mathcal B}$ component on the boundary conditions, and the $\bm{\mathcal D}$ component is the empirical loss on the 
training data.
Specifically, the individual loss components are formulated as follows:
    \begin{align}
        \mathcal L_{\bm{\mathcal F}}(\bm z;\theta)& = \frac{1}{N_F} \sum_{i=1}^{N_F} \| \bm{\mathcal F}(\hat {\bm u}_{\theta} (\bm z_i); \gamma) - \bm f(\bm z_i)\|^2, \\
        \mathcal L_{\bm{\mathcal B}}(\bm z;\theta)& = \frac{1}{N_B} \sum_{i=1}^{N_B} \| \bm{\mathcal B}(\hat {\bm u}_{\theta} (\bm z_i)) - \bm g(\bm z_i)\|^2, \\
        \mathcal L_{\bm{\mathcal D}}(\bm z;\theta)& = \frac{1}{N_d} \sum_{i=1}^{N_d} \| \hat {\bm u}_{\theta} (\bm z_i^*) - \bm u_i^*\|^2,
    \end{align}    
where $N_F$, $N_B$, and $N_d$ are the numbers of differential equations, boundary conditions, and training set points, respectively, and $(\cdot)^*$ denotes ground-truth values. 
When physics-informed loss components $\mathcal L_{\bm{\mathcal F}}$ and $\mathcal L_{\bm{\mathcal B}}$ are not included, Eq.~\eqref{eq:pi_loss_general} simplifies to $\mathcal L(\theta) =  w_{\bm{\mathcal D}} \mathcal L_{\bm{\mathcal D}}(\theta)$, which corresponds to the mean squared error (MSE) loss often used in supervised regression.

\subsection{Mode-Adaptive Neural Networks}
In this work, we adopt Mode-Adaptive Neural Networks (MANN) \cite{Zhang2018} as a supervised learning architecture for trajectory generation.
MANN is well-suited to learning a variety of locomotion modes for legged models.
The MANN input, $\mathbf x_i$, is composed of selected model state information as well as user inputs.
From $\mathbf x_i$, MANN predicts the desired legged model state at the next time step $\mathbf x_{i+1}$.
The MANN architecture consists of 2 neural network (NN) components: a gating network and a motion prediction network.
The gating network employs the Mixture of Experts (MoE) paradigm~\cite{Jacobs1991} and computes a set of $K$ blending coefficients $\bm \theta_i = \{\theta_{i1}, \ldots, \theta_{iK}\}$ in the feature space, given a subset $\hat{\mathbf x}_i$ of the full NN input $\mathbf x_i$.
Then, the motion prediction network  weights $\bm \hat{\alpha}_i=\sum^K_{j=1} \theta_{ij} \bm \alpha_j$ are computed using the predicted blending coefficients, with the $K$ expert weights $\{\bm\alpha_1, \ldots, \bm\alpha_K\}$. 
With these weights and the model state and a user input, the motion prediction network predicts the state at the next time step, $\mathbf y_i$. 
In the original MANN implementation, the training loss function is simply the mean squared error (MSE) between the predictions and the labels in the dataset, i.e., $\mathcal L_{\bm{\mathcal D}}(\bm z;\theta)$.
We remark that prior work employing MANN for locomotion trajectory generation only considered this data-fitting loss term~\cite{Yang2020, Viceconte2022}.

\subsection{Trajectory Controllers}
\label{sec:traj_controllers}
To execute the generated trajectories on the robot, multiple types of controllers can be adopted.
In humanoid robotics, a nested control architecture is typically employed consisting of 3 layers: \emph{trajectory generation}, \emph{trajectory adjustment}, and \emph{trajectory control} \cite{Romualdi2018}.
The \emph{trajectory generation} layer parametrizes the high-level path for the robot to follow in terms of contact, joint, and centroidal dynamics \cite{Romualdi2024}. In our work, the trajectory is generated via our physics-informed BC approach. 

We then stabilize generated trajectories with both an offline and online controller.
This section describes the two control strategies.

\subsubsection{Offline control}
The offline controller makes use of the full trajectory generated by the \emph{trajectory generation} layer. The offline \emph{trajectory adjustment} layer stabilizes
the zero moment point (ZMP) position, center of mass (CoM) position, and CoM velocity over the duration of the trajectory.
The final \emph{whole-body quadratic programming (QP) control} layer solves for the optimal robot velocity over the full trajectory
given a set of tasks. Details of this control scheme can be found in \cite{Romualdi2020}.

\subsubsection{Online control}
Unlike the offline controller, the online controller performs the \emph{trajectory generation} step online during deployment.
the \emph{trajectory adjustment} step is formulated as a model-predictive control (MPC) optimization problem, which stabilizes a trajectory that has been generated at that iteration for the future horizon. The MPC problem
adjusts the centroidal momentum and contacts based on the future horizon prediction from the \emph{trajectory generation} layer as well as constraints placed on the momentum and contacts for stability.
Finally, at each iteration the \emph{trajectory control} layer uses robot feedback and outputs of the previous two layers to track the desired ZMP and CoM, generate swing foot trajectories, and finally compute the joint values with a QP
optimization problem.
Details are present in \cite{Romualdi2024}.

\begin{figure}
    \centering
    \vspace{2mm}\includegraphics[width=\linewidth]{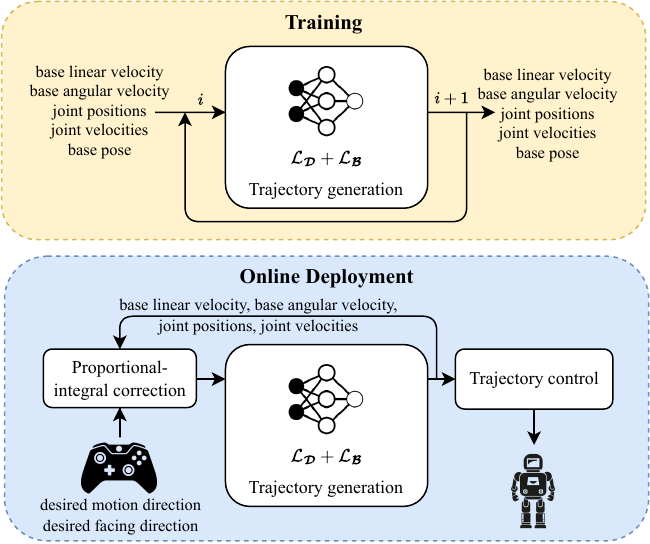}
    \caption{Overall humanoid locomotion architecture integrating the  proposed PI loss and proportional-integral correction block.}
    \label{fig:flowchart}
\end{figure}

\section{METHODS}
\label{sec:methods}

This paper presents two main methodological contributions. 
First, we propose a \emph{physics-informed} architecture for learning the kinematic state of a floating base system subject to constraints.
Second, we propose a \emph{control-informed} correction scheme
which mitigates the drift of the predictions and steers them towards desired values, without the need for retraining or modifying the learning architecture itself.
\Cref{fig:flowchart} summarizes how the two proposed approaches fit into the overall trajectory generation and control scheme which we adopt
to validate their efficacy.

\subsection{Training features}
\label{sec:train_features}
We 
define the input and output features such that the full kinematic state $(\bm q, \bm \nu)$ is learned.
Thus, the input feature vector $\mathbf x_i$ is structured as:
\begin{equation}
    \begin{aligned}
        \mathbf x_i = \{ {}^{\mathcal B} \mathbf V_{\mathcal I, \mathcal B, i-1}, {}^{\mathcal B} \bm \Omega_{\mathcal I, \mathcal B, i-1}, \bm s_{i-1}, \dot{\bm s}_{i-1}, {}^{\mathcal I} \bm p_{\mathcal B,i-1}, {}^{\mathcal I} \bm  \psi_{\mathcal B,i-1} \} \\
        \in \mathbb{R}^{2n + 78},
    \end{aligned}
    \label{eq:nn_input}
\end{equation}
where ${}^{\mathcal B} \mathbf V_{\mathcal I, \mathcal B,i-1}, {}^{\mathcal B} \bm \Omega_{\mathcal I, \mathcal B,i-1} \in \mathbb{R}^{36}$ are the vectorizations of the 12 base linear and angular velocities, expressed in the body-fixed frame, sampled at 6 Hz over a 2 s window centered at the current timestep $i$, and ${{}^{\mathcal I} \bm p_{\mathcal{B},i-1}, {}^{\mathcal I} \bm  \psi_{\mathcal{B},i-1} \in \mathbb{R}^3}$ are the base position and orientation measured in the inertial frame at $i-1$. Similarly, the output vector $\mathbf y_i$ is expressed as:
\begin{equation}
    \mathbf y_i = \{ {}^{\mathcal B} \mathbf V_{\mathcal I, \mathcal{B},i}, {}^{\mathcal B} \bm \Omega_{\mathcal I, \mathcal{B},i}, \bm s_i, \dot{\bm s}_i, {}^{\mathcal I} \bm p_{\mathcal{B},i}, {}^{\mathcal I} \bm  \psi_{\mathcal{B},i} \} \in \mathbb{R}^{2n + 48},
    \label{eq:nn_output}
\end{equation}
where ${}^{\mathcal B} \mathbf V_{\mathcal I, \mathcal{B},i}, {}^{\mathcal B} \bm \Omega_{\mathcal I, \mathcal{B},i} \in \mathbb{R}^{21}$ are the flattened 7 future body-fixed base velocities over a 1 s window of future information, starting at the current timestep, and ${}^{\mathcal I} \bm p_{\mathcal{B},i}, {}^{\mathcal I} \bm  \psi_{\mathcal{B},i} \in \mathbb{R}^3$ contain the base pose information at timestep $i$. 
We provide the full, 3D kinematic state to the MANN model, as opposed to previous work which 
considers only ground projections \cite{Zhang2018, Yang2020, Viceconte2022}.

\subsection{Physics-informed loss component}

We propose a PI loss component 
to reduce contact foot sliding.
This is added 
to the MSE loss (see Eq.~\eqref{eq:pi_loss_general}), which penalizes deviations between predicted and actual robot state. 

We design the PI loss 
to encourage the support foot of the humanoid to remain fixed when in contact with the ground:
\begin{equation}
    {\mathbf v}_{SF} = \mathbf J_{SF} \bm \nu = \begin{bmatrix} \mathbf J_{SF}^{\mathcal B} & \mathbf J_{SF}^{\dot {\bm s}} \end{bmatrix} \begin{bmatrix} {\mathbf v}_{\mathcal B} \\ \dot {\bm s } \end{bmatrix} = 0,
    \label{eq:kin_feas}
\end{equation}
where $SF$ stands for the support foot, while $\mathbf J_{SF}^{\mathcal B} \in \mathbb R^{6 \times 6}$ and $\mathbf J_{SF}^{\dot {\bm s}} \in \mathbb R^{6 \times n}$ indicate the components of the Jacobian that map to the $SF$ velocities from the 6D base velocity and joint velocities, respectively. Eq.~\eqref{eq:kin_feas} can be rearranged to act as a constraint on the velocity of the floating base:
\begin{equation}
    {\mathbf v}_{\mathcal B} = -(\mathbf J_{SF}^{\mathcal B})^{-1} \mathbf J_{SF}^{\dot {\bm s}} \dot {\bm s },
    \label{eq:kf_base}
\end{equation}
where $\mathbf J_{SF}^{\mathcal B}$ is an upper triangular block matrix whose diagonal blocks are rotation matrices, and is thus always invertible.

We formulate the PI loss component as:
\begin{equation}
\begin{aligned}
    \mathcal L_{\bm{\mathcal B}}(\mathbf x;\theta) = \Big\| &\begin{bmatrix} {}^{\mathcal B} \hat{\bm v}_{\mathcal I, \mathcal B} \\ {}^{\mathcal B} \hat{\bm \omega}_{\mathcal I, \mathcal B} \end{bmatrix} (\mathbf x;\theta) + 
    \bigl[ \alpha (\mathbf J_{LF}^{\mathcal B})^{-1} \mathbf J_{LF}^{\dot{\mathbf s}}  + \\
    &(1-\alpha)(\mathbf J_{RF}^{\mathcal B})^{-1} \mathbf J_{RF}^{\dot{\mathbf s}} \bigr] \hat {\dot{\mathbf s}} (\mathbf x;\theta) \Big\| ^2,
\end{aligned}
\label{eq:pi_loss}
\end{equation}
where 
$\mathbf x$ is the input feature vector, and $\alpha$ is a binary variable that is set to $0$ when the right foot (RF) is the support foot, or $1$ when the left foot (LF) is the support foot.

The PI loss term is included in 
the overall loss function $\mathcal{L (\mathbf x;\theta)}$, together with the data-fitting loss $\mathcal L_{\bm{\mathcal D}}$:
\begin{equation}
    \mathcal L (\mathbf x;\theta) = \mathcal L_{\bm{\mathcal D}}(\mathbf x;\theta) + w \mathcal L_{\bm {\mathcal B}}(\mathbf x;\theta),
    \label{eq:overall_loss}
\end{equation}
where $w \equiv w_{\bm{\mathcal B}}$ is the positive scalar weight given to the PI loss function component during training of the network.
We set $w_{\bm{\mathcal D}} = 1$ to simplify parameter tuning.

\begin{figure}[h!]
\centering
\begin{subfigure}{0.48\textwidth}
    \includegraphics[width=\linewidth]{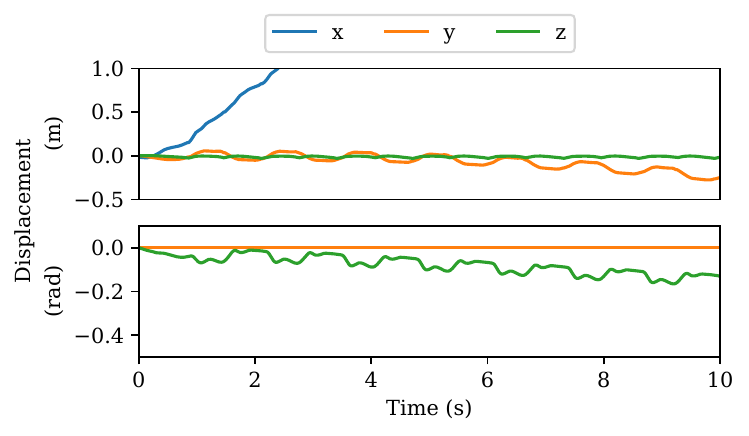}
    \caption{Base displacement for a baseline model \cite{Viceconte2022}. In that formulation, the $x$ and $y$ base rotations were constrained to 0.}
\end{subfigure}
\hfill
\begin{subfigure}{0.48\textwidth}
    \includegraphics[width=\linewidth]{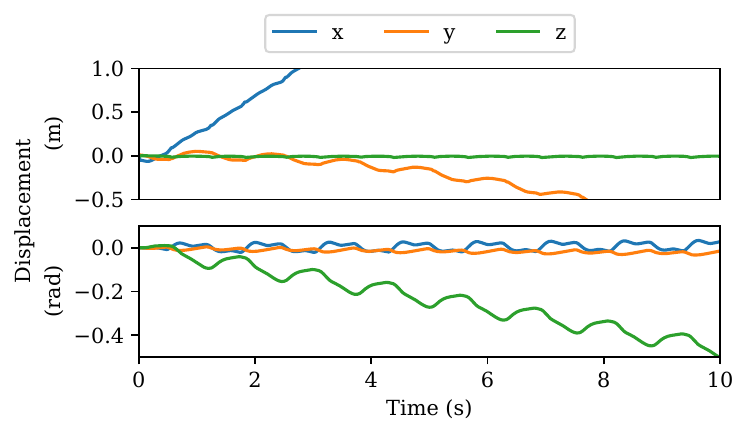}
    \caption{Base displacement when correction block is disabled.}
\end{subfigure}
\hfill
\begin{subfigure}{0.48\textwidth}
    \includegraphics[width=\linewidth]{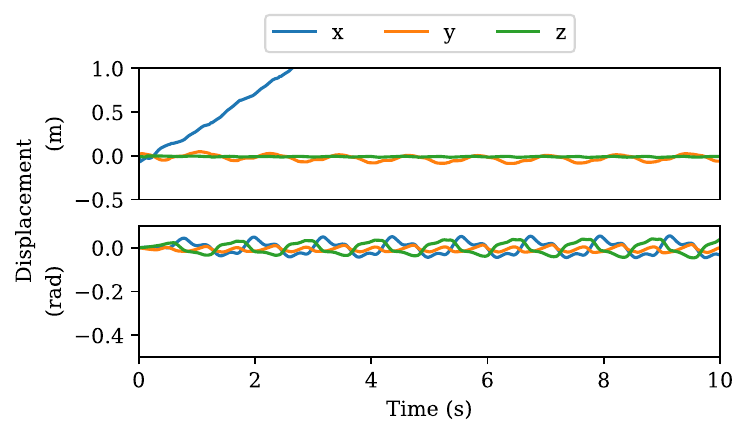}
    \caption{Base displacement when correction block is enabled.}
\end{subfigure}
\caption{Comparison of the drift in base position (top) and orientation (bottom) for (a) baseline trajectory generation, (b) our approach without drift correction block, and (c) our approach with drift correction block for forward walking. All values are measured w.r.t. the inertial frame, and the robot starts at the inertial origin, facing the positive $x$ direction.}
\label{fig:base_displacement_pid_ablation}
\end{figure}

\subsection{Control-informed steering for network drift correction}

One drawback of 
IL methods is that due to biases in training data, a trained model may produce asymmetric outputs and thus cause the trajectory to drift from the desired direction.
Assuming an arbitrary learning model which predicts the kinematic state $(\bm q, \bm \nu)$, such as the one depicted in the top of \Cref{fig:flowchart}, given the previous state, we propose closing the loop on $\bm \nu$ using the desired user input $\bm q_{\mathcal B,d}$.
This solution requires no changes to the training or inference of the learned model, and allows the user specifications to directly steer the trajectory.
This correction block removes drift by counteracting learned biases
and enforces additional desired dynamics into the system to naturally slow down and stop the walking at the goal position and orientation.

Our drift correction is composed of a linear part and an angular part. The linear velocity controller is formulated as:
\begin{equation}
{}^{\mathcal B} \boldsymbol v^c_{\mathcal I, \mathcal B} = {}^{\mathcal B} \hat{\boldsymbol v}_{\mathcal I, \mathcal B} - k_0 \left({}^{\mathcal I} \bm R_{\mathcal B} \right)^{\top} \left({}^{\mathcal I} \bm p_{\mathcal B} - {}^{\mathcal I} \bm p^d_{\mathcal B}\right),
    \label{eq:lin_pid}
\end{equation}
where ${}^{\mathcal B} \boldsymbol v^c_{\mathcal I, \mathcal B}$ denotes the linear base velocity control output for the next network input, ${}^{\mathcal B} \hat{\boldsymbol v}_{\mathcal I, \mathcal B}$ is the network-predicted linear base velocity, $k_0$ is the positional error gain, ${}^{\mathcal I} \bm R_{\mathcal B}$ is the current base orientation, and ${}^{\mathcal I} \bm p_{\mathcal B}$ and ${}^{\mathcal I} \bm p^d_{\mathcal B}$ are the current and desired base positions. 

Instead, the rotational velocity controller is defined based on \cite[Sec. III.E]{Pucci2018} and \cite[Sec. 5.11.6]{Olfati2001}, such that the generated state approaches the desired base rotation ${}^{\mathcal I} \bm R^{d}_{\mathcal B}$:
\begin{equation}
{}^{\mathcal B} \bm \omega^c_{\mathcal I, \mathcal B} = {}^{\mathcal B} \hat{\bm \omega}_{\mathcal I, \mathcal B} - k_1 \mathrm{skew}\left({}^{\mathcal I} \bm R^{d \top}_{{\mathcal B}} {}^{\mathcal I} \bm R_{\mathcal B}\right)^{\vee},
    \label{eq:rot_pid}
\end{equation}
where ${}^{\mathcal B} \bm \omega^c_{\mathcal I, \mathcal B}$ becomes the angular base velocity for the MANN input at the next time step, ${}^{\mathcal B} \hat{\bm \omega}_{\mathcal I, \mathcal B}$ is the predicted angular velocity, $k_1$ is the positional error gain, and $\mathrm{skew}(\cdot)^{\vee}$ is the vectorized skew-symmetric part of a square matrix.

\begin{figure}
\centering
\begin{subfigure}{0.4\textwidth}
    \includegraphics[width=\linewidth]{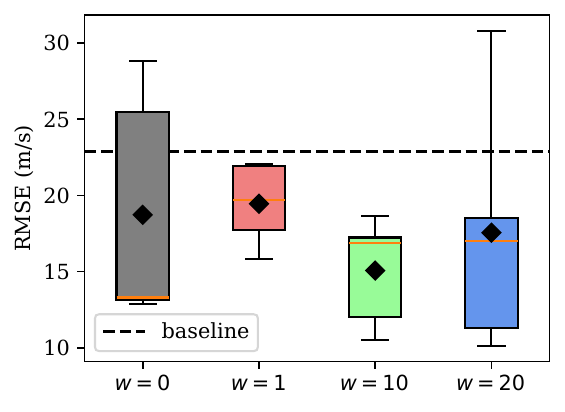}
    \caption{Average foot linear velocity RMSE.}
\end{subfigure}
\hfill
\begin{subfigure}{0.4\textwidth}
    \includegraphics[width=\linewidth]{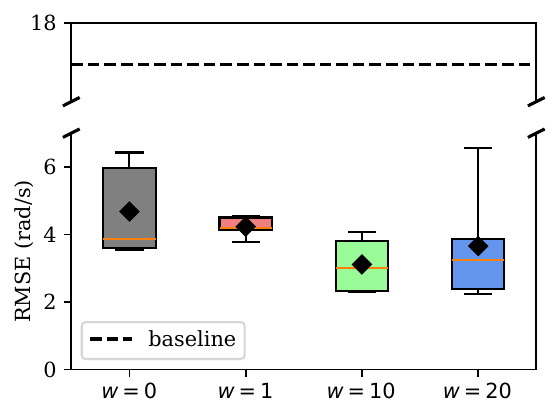}
    \caption{Average foot angular velocity RMSE.}
\end{subfigure}
\caption{Comparison of the average total RMSE of the velocity of the support foot during a 10 s forward walking trajectory, for $w \in \{0, 1, 10, 20\}$ ($w=0$ means $\mathcal L_{\bm{\mathcal D}}$ only). Medians are represented by orange lines and means as black diamonds. Baseline approach is from \cite{Viceconte2022}.}
\label{fig:pi_weight_comparison_plot}
\end{figure}

\section{RESULTS AND DISCUSSION}
\label{sec:results}

For validation of the proposed approach, we train the proposed model on 30 minutes of multi-directional human locomotion data using the XSens~\cite{Roetenberg2009} sensor suit.
The dataset size is doubled by mirroring the base pose across the world $xz$ plane and joint positions across the sagittal plane. 
In total, this corresponds to roughly $200,000$ data points sampled at 50 Hz, which are then retargeted onto the 160 cm, 56 kg ergoCub humanoid robot,
considering $n=26$ joints. We use the same dataset and retargeting method as \cite{Viceconte2022}.

MANN model training is carried out using PyTorch with 4 experts over the course of 150 epochs, with the data downsampled to 50 Hz and making use of vectorized batch computations for higher efficiency.
The single hidden layer of the gating network contains 32 neurons while that of the prediction network has 512.
We use the AdamW optimizer to train the model~\cite{loshchilov2017decoupled}, requiring approximately 24 hours on a single NVIDIA GeForce RTX 3050 Ti GPU.
Experimental code
and datasets are available at: \url{https://github.com/ami-iit/paper_delia_2025_iros_physics-informed_trajectory_generation}.

In order to assess the capabilities of our proposed approach, we perform three
experiments:
i)~a comparison of the generated trajectories with and without our physics-informed loss and correction block, 
ii)~an empirical analysis of our approach on the ergoCub robot combined with both an offline and an online controller, 
and iii) an ablation study isolating the effects of the proposed correction block and the PI loss component on the generated trajectories.

\begin{figure*}
    \vspace{4pt}
    \centering
\begin{subfigure}{0.48\textwidth}
    \includegraphics[width=\linewidth]{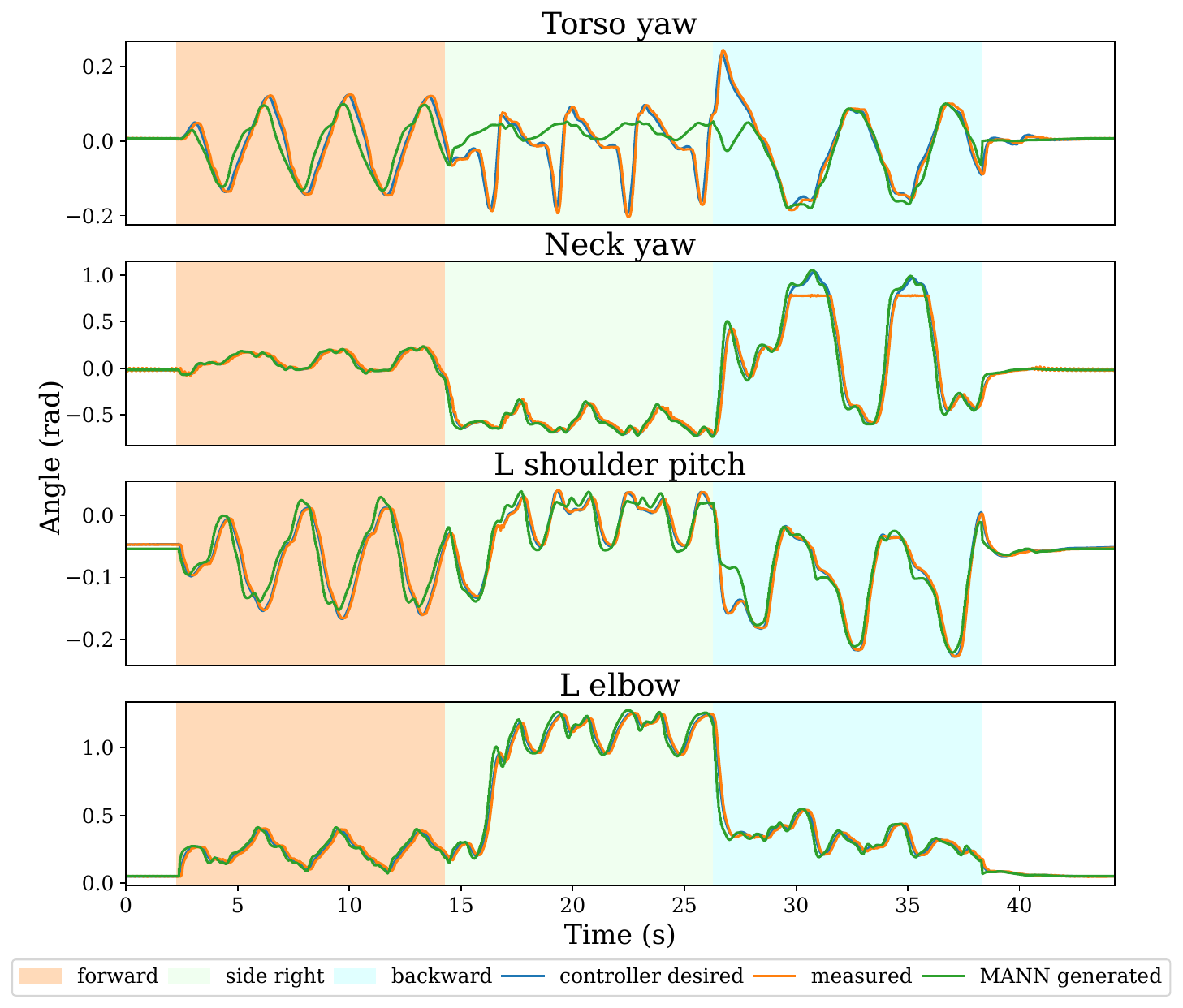}
    \caption{Offline DCM controller results at 0.33x speed and \newline 0.4x step size.}
\end{subfigure}
\begin{subfigure}{0.48\textwidth}
    \includegraphics[width=\linewidth]{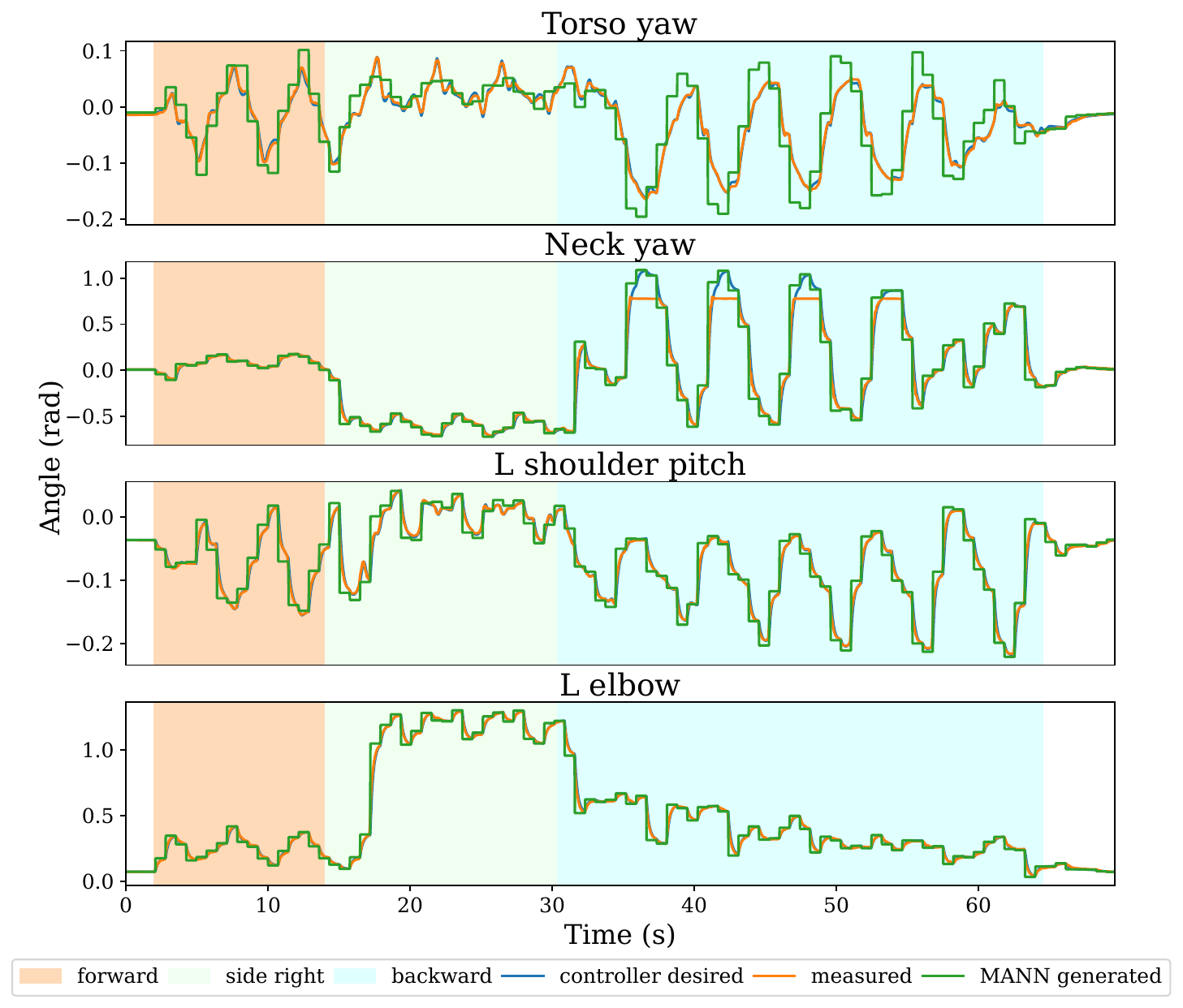}
    \caption{Online MPC controller results at 0.25x speed and \newline 0.3x step size.}
\end{subfigure}
\caption{Side-by-side of selected joint tracking for two different controllers when deployed on the ergoCub robot. Data in each case contains a mixed walking trajectory with smooth transitions: standing, forward, side, backward, and standing again.}
\label{fig:joint_plots}
\end{figure*}

\begin{figure}
    \centering
    \includegraphics[width=\linewidth]{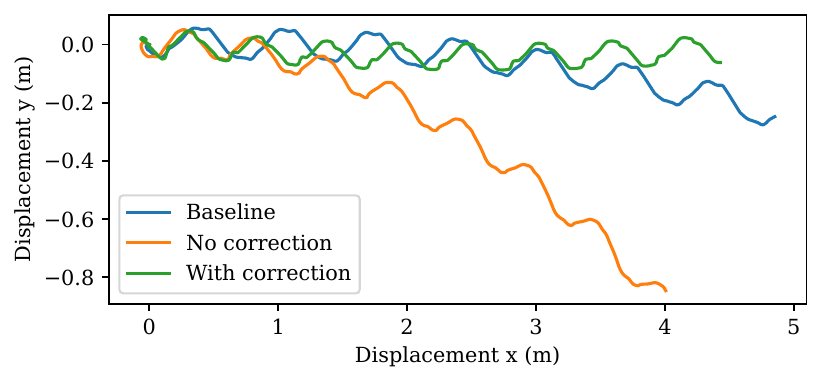}
    \caption{Ground-projected base displacements for the baseline model, our model without the correction, and our model with the correction, for a 10 s forward walking trajectory.}
    \label{fig:ground_projection}
\end{figure}

\subsection{Performance of our approach}
\label{sec:performance}
We compare models trained using our approach to one trained 
using the MANN implementation presented in~\cite{Viceconte2022} as a baseline for comparison.
We observe that our approach produces measurable improvements in the physical feasibility and how effectively the trajectories adhere to user input.
In particular, 
as shown in \Cref{fig:base_displacement_pid_ablation}, 
the deviation from the desired walking direction is significantly reduced between the baseline (a) and our approach with the correction block (c). In particular, we observe the $y$ linear and $z$ angular drift. The $x$ displacement is expected due to the walking direction.

Notably, the drift for our model without a correction block (b) is larger than that of the baseline model (a).
This is because, unlike the baseline, the new model allows for variations in the base roll and pitch, making the model more versatile.
Nonetheless, the drift is successfully counteracted when our proposed correction block is activated.

In terms of foot sliding, models trained with our proposed approach display clear advantages.
In fact, the baseline model
results in average support foot RMSE values of $22.88$~m/s for the linear velocities and $17.74$~rad/s for the angular velocities.
As shown in \Cref{fig:pi_weight_comparison_plot}, such errors are 
significantly larger than those obtained by the proposed approach across the considered PI loss weights 
$w \in \{ 0,1,10,20\}$, 
especially for the angular velocities. Accordingly, reduced support foot sliding translates to better stability during locomotion.

\subsection{Real-world experiments on the ergoCub robot}
To demonstrate the modularity and successful real-world application of our proposed
approach, we deploy our trajectory generator with two different tracking controllers
on the ergoCub humanoid robot.
The two controllers, described in \Cref{sec:traj_controllers}, 
perform slightly differently, yet ultimately track the generated trajectories successfully
as shown in the accompanying video.

\Cref{fig:joint_plots} compares the tracking of generated upper-body joint trajectories for multiple walking directions and transitions between the two controllers.
When testing on the real robot, it is sometimes necessary to scale the footstep size or the trajectory speed to improve dynamic stability. 
As reported in the captions of \Cref{fig:joint_plots}, the offline controller achieves slightly faster walking and slightly larger step sizes.
Furthermore, the joint trajectory tracking of the offline controller is noticeably smoother than that of the online controller.
This may be because the online controller's \emph{trajectory control} block runs at a much lower frequency, which reduces its reaction speed to fast motions, resulting in slightly smaller and slower steps.
An interesting emergent human-like behavior is MANN's generation of backward-facing neck configurations when following a backward walking trajectory. With both control schemes, the controller modifies the MANN output which exceeds the neck yaw limit for backward walking and generates very small torso yaw displacements for side walking.

\begin{figure}
    \centering
        \includegraphics[width=\linewidth]{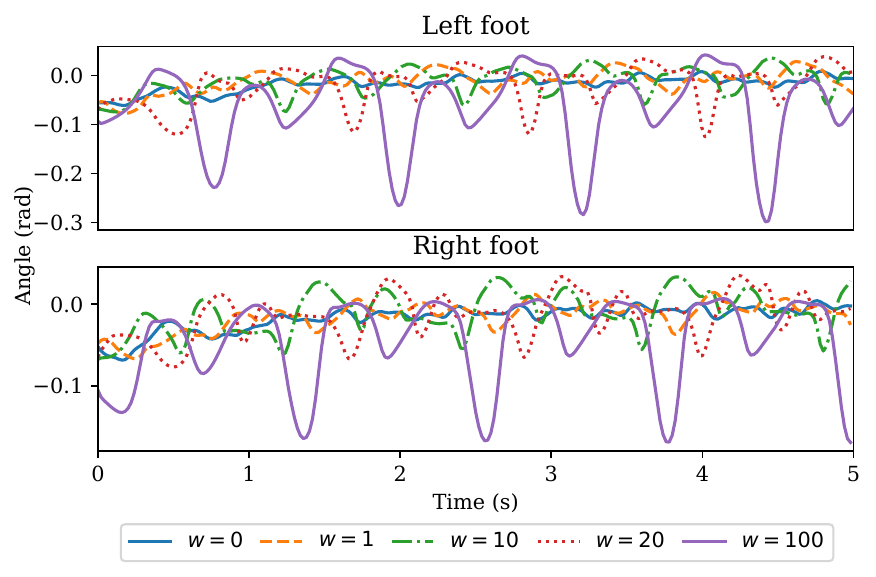}
    \caption{Left and right foot pitch values over time for a forward walking trajectory, based on the PI loss weight used.}
    \label{fig:foot_rotations}
\end{figure}

\subsection{Ablation studies}

With the aim of quantifying
the effects 
of our proposed techniques
on the quality of the generated trajectories, we perform experimental comparisons 
by selectively deactivating either or both the drift correction block and the PI loss component during training.

\subsubsection{Drift correction block}

We compare the base pose trajectory
with and without the correction block to test its efficacy. For a short forward walking trajectory, it is already clear from \Cref{fig:base_displacement_pid_ablation} that the correction block mitigates the drift from the desired forward walking trajectory over 10 s. In particular, the correction eliminates the significant lateral ($y$) and yaw drift from the raw generated trajectory.
\Cref{fig:ground_projection} also shows that the ground-projected base displacement for the same forward trajectory follows the desired walking direction significantly better with the correction block than without.

\subsubsection{PI loss component}

The efficacy of the PI loss component is evaluated
with the drift correction block deactivated,
to exclude the confounding factor of 
gain tuning.

First, we perform a statistical analysis to determine which $w$ results in the highest model performance.
Five different models are trained with each of the following weights: $w=\{0, 1, 10, 20, 100\}$.
For each model, the epoch with the lowest overall test loss is employed to evaluate the amount of sliding of the support foot during a 10-second forward walking trajectory. 
Then, the RMSE of the support foot velocity at any given time is summed over the duration of the trajectory. \Cref{fig:pi_weight_comparison_plot} shows that increasing $w$ up to $10$ results in a decrease in support foot sliding. 
%

Interestingly, without the correction block, the models trained using the largest PI weight, $w=100$, are not able to walk at all.
Instead, with the correction enabled, walking is achieved, which may indicate that the guidance the correction provides helps overcome undesirable learned artifacts.

In fact, we observe that when $w$ is too large, the PI loss component overpowers the $\mathcal L_{\bm{\mathcal D}}$ component, unexpectedly causing excessive foot rotation to arise, which detracts from the smoothness and human-likeness of the trajectories. \Cref{fig:foot_rotations} displays the generated foot pitch over time for models trained with simple data loss and PI weights of 1, 10, 20, and 100. 
We note that the pitch deviations become more substantial as the PI weight increases.
This may be a result of the PI loss only encouraging the foot to have zero velocity when in contact with the ground, and not explicitly encouraging it to be parallel.
If this is the case, the result is that the learned model ``cheats" by keeping only the heel on the ground during the terminal stance phase, probably because this helps to keep the foot stationary during the phase where it is most likely to slip.
Due to the aforementioned tradeoffs, it was determined that training with $w=10$ produces models with the best compromise of minimizing contact foot sliding while also retaining stable walking behavior.

\begin{figure} 
\centering
\begin{subfigure}{0.48\textwidth}
    \includegraphics[width=\linewidth]{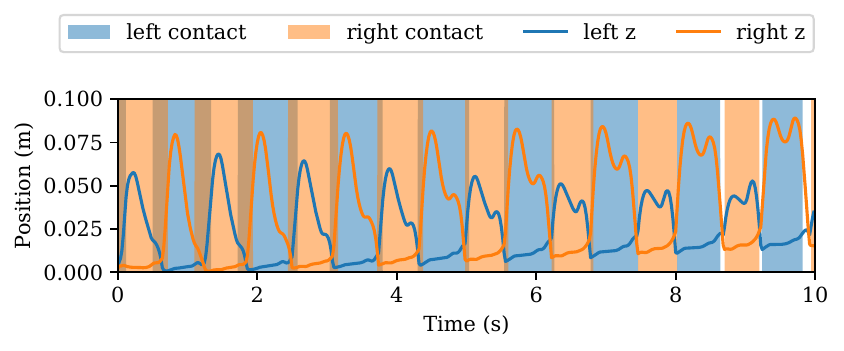}
    \caption{Foot heights from model trained with $\mathcal L_{\bm{\mathcal D}}$ loss.}
\end{subfigure}
\hfill
\begin{subfigure}{0.48\textwidth}
    \includegraphics[width=\linewidth]{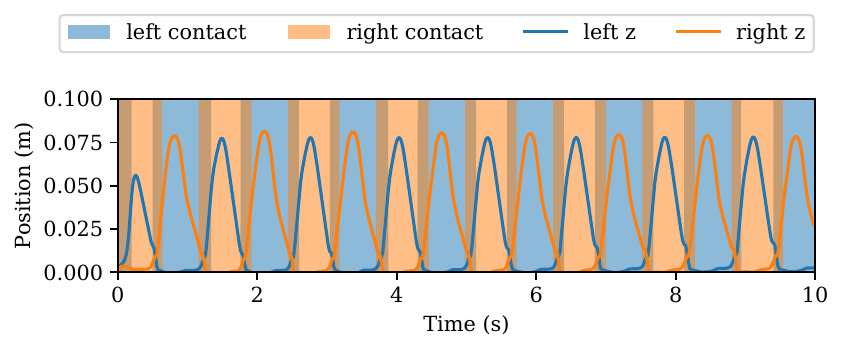}
    \caption{Foot heights from model trained with $\mathcal L_{\bm{\mathcal D}} + 10\mathcal L_{\bm{\mathcal B}}$ loss.}
\end{subfigure}
\caption{Comparison of the variation in foot height over 10 s of forward locomotion. Foot height should remain constant when in contact with the ground.}
\label{fig:pi_ablation_foot_z_plot}
\end{figure}

Given the above results, we show a comparison of the foot height during forward locomotion between a model trained with $\mathcal L_{\bm{\mathcal D}}$ (no PI component) and a model trained with the addition of PI loss with $w=10$.
We can clearly see in \Cref{fig:pi_ablation_foot_z_plot} that 
the PI loss significantly decreases support foot displacement and yields more consistent step heights. 

\section{CONCLUSION AND FUTURE WORK}
\label{sec:conclusion}

This paper presents a method which improves the physical fidelity and input-following of IL humanoid trajectory generation. There are two main contributions, a \emph{physics-informed} architecture and a \emph{control-informed} correction block.
We have shown that these 
generalizable improvements can measurably improve a deep learning approach's adherence to real-world physics laws and desired behavior. Physics-informed learning and 
corrections to learned predictions result in safer locomotion and
more accurate input following 
for a humanoid robot trajectory generation task.
Also, this approach has the potential to be used for many other robotics and machine learning applications 
due to its modularity.

The presented approach only validates one type of PI implementation. It also requires postprocessing of network outputs in order to sample the contacts needed for the other control layers.
Future work will investigate
use
of PIML with more recent IL methods such as diffusion models \cite{Huang2025, Serifi2024}, and other physics priors, such as ground reaction forces or dynamic stability, in the  loss function.


\bibliographystyle{IEEEtran}
\bibliography{bib}



\end{document}